# Stochastic LLMs *do not* Understand Language: Towards Symbolic, Explainable and Ontologically Based LLMs

Walid S. Saba

Institute for Experiential AI, Northeastern University, Portland ME 08544, USA
`w.saba@northeastern.edu`

**Abstract.** In our opinion the exuberance surrounding the relative success of data-driven large language models (LLMs) is slightly misguided and for several reasons (i) LLMs cannot be relied upon for factual information since for LLMs all ingested text (factual or non-factual) was created equal; (ii) due to their subsymbolic nature, whatever 'knowledge' these models acquire about language will always be buried in billions of microfeatures (weights), none of which is meaningful on its own; and (iii) LLMs will often fail to make the correct inferences in several linguistic contexts (e.g., nominal compounds, copredication, quantifier scope ambiguities, intensional contexts. Since we believe the relative success of data-driven large language models (LLMs) is not a reflection on the symbolic vs. subsymbolic debate but a reflection on applying the successful strategy of a bottom-up reverse engineering of language at scale, we suggest in this paper applying the effective bottom-up strategy in a symbolic setting resulting in symbolic, explainable, and ontologically grounded language models.

## 1 Introduction

The recent successes of so-called large language models (LLMs) have taken the world of artificial intelligence (AI) and natural language processing (NLP) by storm. Indeed, with the release of GPT-4 it has become apparent that large language models (LLMs), that are essentially a massive experiment in a bottom-up reverse engineering of language, have crossed some threshold of scale at which point there was an obvious qualitative improvement in their capabilities[1]. In our opinion, however, the spectacular exuberance towards these advances is slightly misguided. For one thing, these large 'language models' are not exactly models of language but are statistical models of regularities found in language. In fact, and due to their subsymbolic nature, whatever 'knowledge' these models acquire about how language works will always be buried in billions of microfeatures (weights) none of which is meaningful on its own. This is also the reason why *explainability* can never be attained in such models since explainability is 'inference in reverse' and this can only happen in one of two ways: (i) either the computation is invertible, or (ii) symbolic structures must be employed to perverse a *semantic map* of the computation, and neither of this is true in deep neural networks (DNNs). As shown in figure 1(a) computation in DNNs proceeds by performing a linear combination (a weighted some) of the inputs followed by the application of some activation function. But this computation is not invertible, since the decomposition of a tensor (vector, or scalar) is undecidable. Thus, once a computation is performed in the forward path of a DNN the constituents of the compositional computation are not anymore available, unlike computations in symbolic systems where structures such as abstract syntax trees are used to preserve a semantic map of the computation. Note that the inputs $x_1$ and $x_2$ (left leaf nodes) and weights $w_1$ and $w_2$ (right leaf nodes) in figure 1(b) are just one combination of infinitely many that produces 0.87 as an output.[2]

---

[1] GPT stands for 'Generative Pre-trained Transformer', an architecture that OpenAI built on top of the transformer architecture introduced in (Vaswani, et. al., 2017).

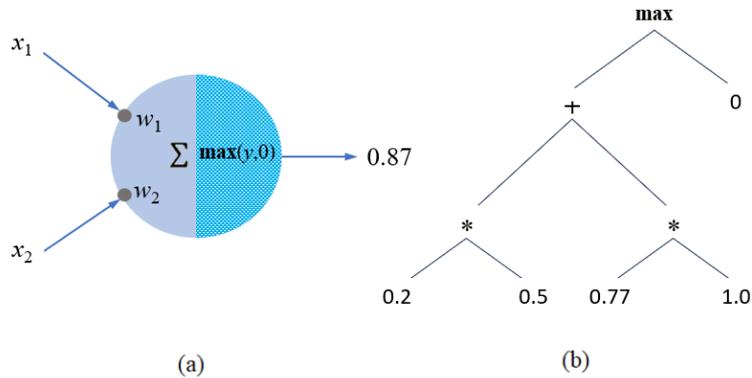

(a)                              (b)

**Figure 1.** Neural (subsymbolic) computations are not invertible: in (a) we cannot explain how we arrived at 0.87; in symbolic systems, on the other hand, a semantic map of the computation is maintained, for example using an abstract syntax tree as shown in (b).

In addition to the issue of *explainability*, (i) LLMs will always be susceptible to bias and toxicity in their training data; (ii) LLMs can never differentiate factual information from non-factual information since for LLMs all text was created equal; and (iii) LLMs will fail to capture the inferential aspects of linguistic communication in many contexts. It is the last point that we would like to focus on here since this is what will hinder the use of these stochastic models in the enterprise, and in particular in critical applications such as finance, the judiciary system, healthcare, etc.

In the rest of the paper we will first examine some linguistic contexts where, regardless of scale and the specific details of the neural model, LLMs will always fail to make the correct inferences. Subsequently, we will briefly discuss why LLMs got farther than symbolic top-down approaches and how the same bottom-up reverse engineering strategy could be done in a *symbolic* setting. We will then briefly discuss how a symbolic bottom-up reverse engineering process also allows us to 'discover' the nature of the ontology that seems to underlie our ordinary spoken language. Finally, we will show how discovering the symbolic dimensions of meaning along with the underlying ontology can help us deal with well-known challenges in the semantics of natural language.

---

[2] See (Saba, 2022) for a more detailed discussion on the relationship between compositionality, structured semantics and explainability, and (Fodor and Pylyshyn, 1988) for a more detailed critic of subsymbolic systems and their inadequacy in preserving semantic systematicity.

## 2   Language 'Understanding' and the Limitations of LLMs

Claims about the linguistic competency of natural language understanding (NLU) systems are often exaggerated (positively or negatively), ranging from claims of human-like performance to systemic inadequacy (Sugawara and Tsugita, 2023). In our opinion this is due to overlooking the difference between subjective and objective evaluations of NLU systems. For example, it is difficult to come up with an objective measure by which one can evaluate an NLU system's *text summarization*. The same is true of the results of a *search* query, or of an extracted set of *key topics* (see figure 2a).

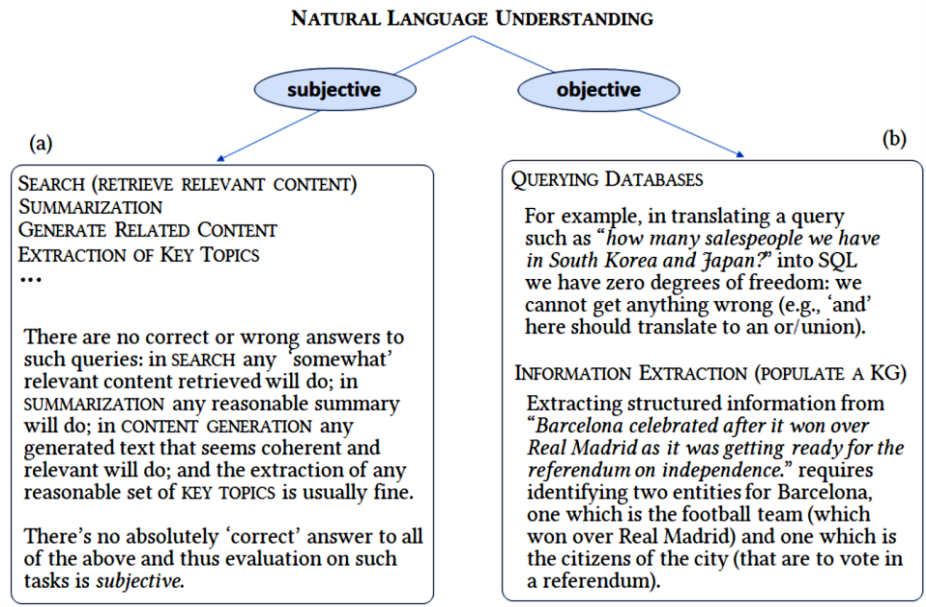

**Fig. 2.** Tasks that can be objectively evaluated vs. tasks that can only be subjectively evaluated.

While most of what are referred to as 'downstream NLP tasks' can in fact only be subjectively evaluated, objective measures do exist for some other NLP tasks; for example the translation of a natural language query into a formal SQL query to be executed against some enterprise database. In translating a query such as '*how many salespeople in the Chicago branch do we have now on vacation*' into SQL we have zero degrees of freedom since getting anything wrong will mean getting the wrong answer. The reverse operation, namely translating free-from text into structured information also has zero degrees of freedom (see figure 2b). Successful translation of a natural language question to the correct SQL query or extracting the right entities and relations from free-form text to some relational structure are therefore the right objective tests for language understanding. It is in these situations that an NLU system must *get everything right*, and that means handling a number of linguistic phenomena such as compound nominals, copredication, quantifier scope ambiguities, prepositional phrase attachments, and so on. As discussed below, and notwithstanding their relative success in subjective tasks, LLMs often fail in making the correct inference in these contexts.

## 2.1 Compound Nominals

One linguistic phenomenon that seems to be beyond LLMs is that of nominal compounds in utterances of the form [$Noun_1$ $R$ [$Adj$ $Noun_2$]] where the LLM always picks out $Noun_2$ as the object of modification by $Adj$, although the object of modification could, in general, be $Noun_1$, $Noun_2$, both (where the utterance is considered ambiguous), or neither (where the utterance is simply not sensible)[3]. Here's a simple example:

**PROMPT**   What is 'informed' modifying here?
              Mary gave a very informed talk to the Cognitive Science Group.

**LLM**       In the sentence "Mary gave a very informed talk to the Cognitive Science
              Group," the word 'informed' is modifying the noun 'talk.' This means it's
              describing the type of talk that Mary gave as being an informed talk.

The LLM answer is clearly wrong since it is Mary who was very informed and not the 'talk' itself (which is an abstract event that has a time, a place, and a duration). The same failure occurs when the LLM inferred that the gift (an artifact) is what is 'thoughtful' in "Mary appreciated the thoughtful gift she received from David", while clearly the correct interpretation is that "it was thoughtful of David" to give Mary a gift. There are two other combinations to be tested: (i) tests where the *Adj* in [$Noun_1$ $R$ [$Adj$ $Noun_2$]] could apply to both $Noun_1$ and $Noun_2$, and (ii) tests where *Adj* does not apply to any of the two nouns. In the former case we have an ambiguous utterance (e.g., "Everyone thinks Olga is a beautiful dancer" where 'beautiful' could be modifying Olga as a person, or Olga's dancing). In the second case we have a situation where modifying any of the nouns should be rejected, rendering the whole utterance as not being sensible, as in "The room needs an articulate corner table". Again, the LLMs failed all these combinations, by always deciding that the modified noun is the one that is preceded syntactically by the adjective, even when it was nonsensical. What these tests show is that LLMs do not exhibit the kind of understanding that requires commonsense knowledge, but that they simply make inferences based on statistically significant syntactic patterns.

## 2.2 Intension

LLMs are based on a purely extensional paradigm, namely that of neural networks that can only deal with numerical values (or vectors and matrices of those). But quite a bit of reasoning in language (and cognition, in general) does not happen at the data level

---

[3] For more on nominal compounds see (McShane et. al., 2014) and (Larson, 1998).

but at the conceptual and intensional levels (see von Fintel and Heim, 2002 for a good tutorial on intension in semantics). This might seem like a trivial stumbling block, but far from it. To appreciate this point let me start by showing an example where, regardless of what our high school teachers taught us, we cannot assume that SQRT(256) and 16 are replaceable in all contexts. Consider the following:

**PROMPT**  Since SQRT(256) = 16, can I replace one for the other in "I saw Mary teaching her little brother that 7 + 9 = 16"?

**LLM**  Yes, you can replace the square root of 256 with its actual value 16 in "I saw Mary teaching her little brother that 7 + 9 = 16." The statement would still be accurate: "I saw Mary teaching her little brother that 7 + 9 = SQRT(256)"

Clearly the answer of the LLM is wrong: from a sentence that reflects reality (it is a fact that "I saw Mary teaching her little brother that 7+9 = 16") we cannot alter or create a new reality and infer something that is not true, namely that "I saw Mary teaching her little brother that 7+9 is SQRT(256)". What's at issue here is that 16 and SQRT(256) are only extensionally equal (i.e., equal by value only) but otherwise they are two different objects. This failure applies not only to mathematical expressions, but to all objects of cognition (anything that we can refer to in language). Consider the following:

**PROMPT**  Since 'Madrid' = 'the capital of Spain' can I replace one for the other in "John thinks Madrid was not always the capital of Spain"

**LLM**  Yes, you can replace 'Madrid' by 'the capital of Spain' in "John thinks Madrid was not always the capital of Spain." The new sentence will be "John thinks Madrid was not always Madrid."

Again, while it is true that Madrid *is* the capital of Spain, the expressions "Madrid" and "the capital of Spain" have different cognitive content and thus the replacement of one by the other in some contexts will lead to wrong (and absurd) conclusions. DNNs (and thus LLMs) cannot account for or model the intension of objects of cognition and thus will always fail in making the right inferences in these contexts.

### 2.3   Copredication

Copredication refers to the situation where a single entity is (implicitly) assigned two different categories (or types) in the same linguistic context (see Asher, 2011; Asher & Pustejovsky, 2006; Viebahn, 2020). For example, in the sentence "That massive red book on the top shelf is written by Tolstoy" we are using a 'book' to refer to two different types of entities, simultaneously: 'book' the **physical** object that is massive and red; and 'book' the **information content** that was authored by Tolstoy. Failure to recognize these copredications results in absurd conclusions as the following illustrates:

**PROMPT**  Suppose this is true:
The wet newspaper that fell down off the table is my favorite newspaper. But now that my favorite newspaper fired the editor I might not like reading it anymore. Can I replace 'my favorite newspaper' by 'the wet newspaper that fell down off the table' in the second sentence?

**LLM**     Yes, you can replace "my favorite newspaper" with "the wet newspaper that fell down off the table" in the second sentence to maintain the same meaning. So the revised sentence would be: "But now that the wet newspaper that fell down off the table fired the editor, I might not like reading it anymore."

The answer the LLM provided is clearly wrong since the **physical object** that is the 'wet newspaper that fall off the table' is not the one that fired 'my favorite columnist' but the organization – or, more accurately, it is a group of people in that organization. The LLM failed here because it failed to detect a *copredication*, namely referring to two types of objects using a single entity. The above example can also be restated so that the erroneous inference is translated into making the wrong reference resolution. Consider a slight change in the above test:

**PROMPT**     What does 'it' refer to here?
The newspaper on the table in front of me used to be my favorite. However, since it fired my favorite columnist, I do not read it anymore.

**LLM**     In the sentence, "it" refers to "the newspaper on the table in front of me." The speaker is talking about a specific newspaper that was their favorite but is no longer because it fired their favorite columnist."

Here the wrong understanding is more explicit by resolving "it" with "the newspaper on the table in front of me" which is clearly wrong since the (physical) newspaper on the table is not the entity that fired 'my favorite columnist'. The failure of LLMs in detecting copredication can even be shown more explicitly. Consider the following test where we explicitly ask GPT to identify and list all entities in some linguistic context, along with their category/type:

**PROMPT**     List all entities mentioned in this sentence, along with their category, or type: The city of Barcelona celebrated on Sunday after it won over Real Madrid as it was getting ready to vote for independence the next day.

**LLM**     Entities mentioned in the sentence:

1.   The city of Barcelona (Category: Geographic Location)
2.   Sunday (Category: Day of the Week)
3.   Real Madrid (Category: Sports Team or Organization)
4.   Independence (Category: Political Action or Event)

These entities encompass a geographical location, days of the week, sports teams, and a political event".

Besides missing "the vote" as the main event, the LLM failed to detect reference to two types of entities that are also implied by 'Barcelona': the Barcelona Football Club (that won over Real Madrid), and *the citizens* of Barcelona (that celebrated the victory and are getting ready to vote for independence).

## 2.4    Quantifier Scope Ambiguity

For decades, the problem of resolving quantifier scope ambiguities has been one of the most challenging problems in computational linguistics (see Kiss and Pafel, 2017).

While quantifier scope is not very important when generating a summary for a piece of text (a subjective NLP task), when translating a question posed in natural language into an unambiguous SQL query deciding on the right scope of quantifiers makes all the difference in getting the correct answer to the query. Since resolving quantifier scope ambiguities often requires making inferences that require some commonsense knowledge (Saba, 2001), LLMs also failed most of these tests. Consider the following:

**PROMPT** How many states are we referring to here?
Every senator in a key electoral state voted for the amendment.

**LLM** In the sentence, "Every senator on a key electoral state voted for the amendment," it implies that there is one key electoral state being referred to. The use of "a key electoral state" suggests a singular state rather than multiple states.

The LLM interpreted the sentence incorrectly as "in some key electoral state, every senator voted for the amendment" which is clearly not the intended meaning. Basically, it would seem that the indefinite "a" triggered a single state, although the scope ordering most speakers would choose is ∀senator∃state, implying several electoral states. The wrong inference is also made here, although the situation is reversed:

**PROMPT** How many journalists are we referring to here?
A journalist from the BBC is always present in every civil war.

**LLM** In the sentence, "A journalist from the BBC is always present in every civil war," it refers to one journalist from the BBC. The use of "a" indicates a singular journalist, not multiple journalists.

The LLM also failed here, although most readers would reverse the scope of quantifiers from ∃j∀w to ∀w∃j thus interpreting the sentence as saying "in every civil war, a journalist from BBC is always present". These failures can be even more acute in situations where some subtle social background knowledge is required, as in the following:

**PROMPT** In the following, are we talking about a specific cowboy or some indefinite cowboy? Next week Susan is going to be married to some cowboy from Arizona.

**LLM** In the sentence, "Susan is going to be married to some cowboy from Arizona," it appears that we are talking about an indefinite cowboy. The use of "some cowboy" suggests that the cowboy is not a specific individual but rather any cowboy from Arizona.

Although it is plausible to assume an indefinite cowboy in "***Someday*** Susan is going to be married to some cowboy from Arizona", but in the world we live in it is unlikely for Susan to be planning on getting married ***next week*** to someone yet to be identified, but that she has *someone specific* in mind. This kind of background knowledge is not available to LLMs, and thus such inferences are clearly beyond these statistical models.

## 2.5 Prepositional Phrase Attachments

Resolving the ambiguity in prepositional phrase attachments is also a very challenging task. Basically, the issue here is what do various prepositional phrases (PPs) attach to (or modify)? While in many situations the attachment is very clear, there are certain situations that require some subtle reasoning that involves access to some background knowledge. Consider for example the following:

**PROMPT**  What does the phrase "in the briefcase" modify here?
Put the book on the table in the briefcase.

**LLM**  In the sentence, "Put the book on the table in the briefcase," the phrase "in the briefcase" modifies "the table." It specifies the location of the table, indicating that the table is inside the briefcase.

The LLM incorrectly read the sentence as [Put the book on [the table in the briefcase]]. Clearly, what is meant here is [Put [the book on the table] in the briefcase] – that is, the book (that is on the table) is to be put in the briefcase and thus the correct attachment of the PP "in the briefcase" is to the verb 'put'. The LLM made the wrong inference here since it has no knowledge of the relation that might exist in the world we live in between books, tables, and briefcases. The LLM also failed in the example below but in this case by incorrectly attaching the PP to the verb[4]:

**PROMPT**  What does "in the U-Haul truck" modify here?
The gangsters decided to sell the shotguns in the U-Haul truck.

**LLM**  "In the U-Haul truck" modifies the verb "sell." It specifies where the gangsters have decided to sell the shotguns, indicating that they plan to conduct the sale from or within the U-Haul truck.

Again, the LLM made the wrong inference since it does not have the relevant background knowledge that it is much more likely for the shotguns to be in the U-Haul truck than for the 'selling of the guns' to occur inside the U-Haul truck.

## 2.6   Metonymy

We conclude examining the linguistic competency of LLMs with another phenomenon in the semantics of natural language, namely metonymy, which is the use of one entity to make an indirect reference to another *related* entity[5]. Consider the following:

**PROMPT**  The waiter told Rita that the loud omelet wants another beer.
Who is it that wants another beer?

**LLM**  In the sentence, "The waiter told Rita that the loud omelet wants another beer," it seems there might be an error or confusion in the wording. Typically, omelets don't have desires or drink beer, so this sentence appears to be constructed in a way that doesn't make logical sense. Please provide additional context or clarification if there's an intended meaning you'd like to convey.

While all competent speakers would immediately infer that "the loud omelet" is a reference to some "loud person eating the omelet", the LLM cannot "uncover" this missing information since it does not 'know' that *eating* is the most salient relationship between people and food (e.g. omelet), although there are many other plausible relations.

---

[4] Example taken from (Peckenpaugh, 2019), with some modification.
[5] See (Shelestiuk, 2005) and (Piñango, et. al., 2017) for a good discussion of metonymy.

### 2.7 Summary of Tests

So where do we stand now? On one hand, LLMs have clearly proven that one can get a handle on syntax and quite a bit of semantics by executing a bottom-up reverse engineering strategy; yet on the other hand what we have are stochastic LLMs that fail to make the correct inferences in many linguistic contexts, and in particular in contexts where some background knowledge might be required. As we believe that the relative success of LLMs is not a reflection on the symbolic vs. subsymbolic debate but is a reflection on (appropriately) adopting a bottom-up reverse engineering strategy, we think that combining the advantages of symbolic representations with a bottom-up reverse engineering strategy is a worthwhile effort. The idea here is that discovering the syntax and semantics of language by analyzing how we actually communicate in language is not exclusive to linguistic work in the empirical tradition but can in fact be done in a symbolic setting. We discuss this next.

## 3 Concerning 'the Company a Word Keeps'

The idea that the meaning of words are acquired by children by observing their syntactic distribution in language is not new (Lidz, 2022). In fact, the genesis of modern LLMs is the *distributional semantics hypothesis* which states that the more semantically similar words are, the more they tend to occur in similar contexts – or similarity in meaning is similarity in linguistic distribution (Harris, 1954). Summarized by "you shall know a word by the company it keeps", a saying attributed to the linguist John R. Firth, the basic idea was to approximate word meanings by embeddings (vectors) that are essentially points in a multidimensional space. While these vectors were initially constructed to approximate word meanings, it was not until the transformer model (Vaswani, et. al., 2017), however, that embeddings started the encoding of *syntactic patterns* and even quite a bit of compositional semantics. At the root of this encoding scheme is a bottom-up approach that "reverse engineers the process and induces semantic representations from contexts of use" (Boleda, 2020). Below we suggest applying this successful bottom-up reverse engineering strategy in a symbolic setting where 'the company a word keeps' is not determined statistically but semantically and ontologically.

### 3.1 Symbolic Reverse Engineering of Language

In discussing possible models (or theories) of the world that can be employed in computational linguistics Jerry Hobbs (1985) once suggested that there are two alternatives: on one extreme we could attempt building a "correct" theory that would entail a full description of the world, something that would involve quantum physics and all the sciences; on the other hand, we could have a promiscuous model of the world that is isomorphic to the way we talk it about in natural language. What Hobbs is suggesting here is a reverse engineering of language to discover how we actually use language to talk about the world. In essence, this is similar to Frege's Context Principal to "never ask for the meaning of words in isolation" (Dummett, 1981) but that a word gets its meanings from analyzing all the contexts in which the word can appear (Milne, 1986).

Again, what this suggests is that the meaning of words can be *discovered* by analyzing all the ways we use these words in everyday discourse. While Hobbs' and Frege's observations might be a bit vague, the proposal put forth by Fred Sommers (1963) is very specific. For Sommers, the process of understanding the meaning of some word *w* starts by enumerating all the properties *P* that can sensibly be said of *w*. For example, while [*delicious apple*] is sensible, [*delicious Thursday*] is not. Moreover, since [*delicious cake*] and [*delicious soup*] are also sensible, there must be a common type (perhaps **food**?) that subsumes *apple*, *soup* and *cake*. Similarly, while [*imminent sugar*] is not sensible, [*imminent trip*], [*imminent conference*] and [*imminent election*] are, again suggesting that *trip*, *conference*, and *election* must have a common supertype (**event**?). Thus, as argued in (Saba, 2007), this type of analysis can also be used to 'discover' the ontology that seems to be implicit in all natural languages.

Let us now consider the following naïve procedure for some initial reverse engineering of language, where **app**(*p*, *c*) means the property *p* can sensibly be said of the concept *c*:

1. Consider concepts $C = \{c_1, ..., c_m\}$ and properties $P = \{p_1, ..., p_n\}$.
2. Assume a predicate **app**(*p*, *c*) that is true iff the property *p* applies to (or can sensible be said of) objects of type *c*, where $c \in C$ and $p \in P$.
3. A set $Cp = \{c \mid \mathbf{app}(p, c)\}$ is generated for all $c \in C$ and all property $p \in P$ such that the property *p* is applicable to (or can sensibly be said of) *c*.
4. A concept hierarchy is then systematically discovered by analyzing the subset relationship between the various sets generated.

Applying the above procedure on a fragment of natural language and taking, initially, *C* to be a set of nouns and *P* a set of adjectives and relations that can sensibly be applied to (or can be said of) nouns in *C*, would result in something like the following:

| | |
|---|---|
| $R_1$ : **app**(OLD, entity) | in ordinary language we can say OLD of any entity |
| $R_2$ : **app**(HEAVY, physical) | we say HEAVY of objects that are of type physical |
| $R_3$ : **app**(HUNGRY, living) | HUNGRY is said of objects that are of type living |
| $R_4$ : **app**(ARTICULATE, human) | ARTICULATE is said of objects that are of type human |
| $R_5$ : **app**(MAKE(human, artifact)) | MAKE holds between a human and an artifact |
| $R_6$ : **app**(MANUFACTURE(human, tool)) | MANUFACTURE relates a human and a tool |
| $R_7$ : **app**(RIDE(human, vehicle)) | RIDE holds between a human and a vehicle |
| $R_8$ : **app**(DRIVE(human, car)) | DRIVE holds between a human and a car |

Note that since **app**(HEAVY, car) – that is, since it is sensible to say 'heavy car' it would seem that car must be a subtype of physical since HEAVY can sensibly be said of all physical things. Similarly, since it makes sense to say MAKE and MANUFACTURE of a tool, a tool must be a subtype of artifact. The fragment hierarchy that is implicit in $R_1$ through $R_8$ is shown in figure 3 below.

## 3.2    Discovering the Language Agnostic Primitive Relations

The fact that **app**(ARTICULATE, human) − namely that it is sensible to say ʹarticulate humanʹ in ordinary discourse, can be restated as **hasProp**(articulation, human), to say that in ordinary discourse it is sensible to attribute the property of articulation to a human. What we have done here is a reification (nominalization) of ARTICULATE to get the abstract object (or trope) articulation (see Moltmann, 2013). The same can be done with **app**(HUNGRY, living) which states that it is sensible to say 'hungry' of any living thing, by restating this as **inState**(hunger, living), which says that is sensible to say that any living thing can be in a state of hunger.

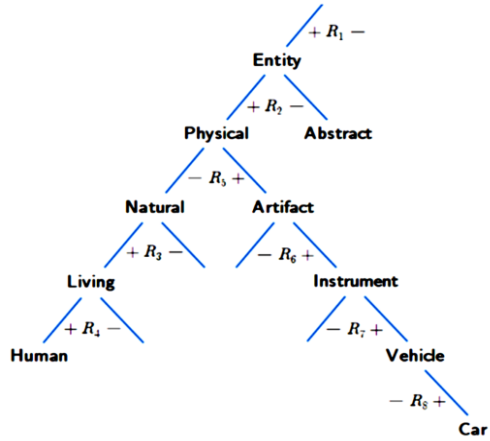

**Fig. 3.** The hierarchy that is implicit in the 'discoveries' $R_1$ through $R_8$ above.

It should be noted here that these transformations add quite a bit of information. While **app**(ARTICULATE, human) and **app**(HUNGRY, living) simply state that it is sensible to say 'articulate' of any human and 'hungry' of any living thing, **hasProp**(articulation, human) and **inState**(hunger, living) are saying that a human can have the *property* of articulation, and that a living thing can be in a *state* of hunger. Other universals can also be obtained. For example, **app**(MANUFACTURE(human, tool)) is saying more than "it is sensible to speak of a human manufacturing a tool" — it is also saying **agentOf**(manufacturing, human) and **objectOf**(manufacturing, tool); i.e., that a human can be the *agent* of a manufacturing (activity) and that tool can be the *object* of such an activity.

These primitive relations (**hasProp**, **inState**, **agentOf**, **objectOf**, etc.) can actually be discovered by analyzing all the ways we describe objects in everyday discourse. In general, when describing an object or an entity $x$ by some property $P$ we are, indirectly, making a statement such as '$x$ is $P$'. If we analyze the various ways these descriptions can be made (using the copular 'is'), it will lead us to different types of primitive relations, as shown in table 1. For example, in saying *Mary is wise*, we are essentially saying that *Mary has the property of wisdom*. Similarly, in saying *Carlos is ill*, we are essentially saying that *Carlos is in the* (physiological) *state of illness*. Analyzing all the ways different types of entities can be described leads us to discover the language agnostic primitive relations summarized in table 2 below.

Here's a summary of the overall process we discussed so far: (*i*) analyze a large corpus to discover facts such as $R_3 :$ **app**(HUNGRY, living) and $R_4 :$ **app**(ARTICULATE, human); (*ii*) construct the ontology implicit in all the discovered relations; and (*iii*) via a nominalization process convert all **app**($p, c$) to two entities related by a primitive, language agnostic relation: **inState**(human, illness) and **hasProp**(human, articulation).

**Table 1.** Discovering primitive relations by analyzing all the ways we can say '*x* **is** *P*'.

| LINGUISTIC CONTEXT | IMPLICIT PRIMITIVE RELATION |
| --- | --- |
| Frido **is** a dog | Frido **instanceOf** dog |
| Billy the Kid **is** William H. Boney | Billy the Kid **eq** William H. Boney |
| JFK **is** John Fitzgerald Kennedy | JFK **eq** John Fitzgerald Kennedy |
| Mary **is** wise | Mary **hasProp** wisdom |
| Julie **is** articulate | Julie **hasProp** articulation |
| Jim **is** sad | Jim **inState** sadness |
| Carlos **is** ill | Carlos **inState** illness |
| Sara **is** running | Sara **agentOf** running |
| Olga **is** dancing | John **agentOf** dancing |
| Sara **is** greeted | Sara **objectOf** greeting |
| Sara **is** acknowledged | Sara **objectOf** acknowledgment |
| John **is** 5'10'' tall | John's *height* **hasValue** 5'10'' |
| Dan **is** 69 years old | Dan's *age* **hasValue** 69 yrs |
| Sheba **is** running | Sheba **participantIn** running (event) |
| Olga **is** dancing | Olga **agentOf** dancing (activity) |

**Table 2.** Some of the language-agnostic primitive relations.

| PRIMITIVE RELATIONS | DESCRIPTION |
| --- | --- |
| **Eq**$(x, y)$ | **individual** $x$ is identical to individual $y$ |
| **Part**$(x, y)$ | individual $x$ is part of individual $y$ |
| **Inst**$(x, y)$ | individual $x$ instantiates **universal** $y$ |
| **Inhere**$(x, y)$ | individual $x$ inheres in individual $y$ |
| **Exemp**$(x, y)$ | individual $x$ exemplifies property $y$ |
| **Dep**$(x, y)$ | individual $x$ depends for its existence on individual $y$ |
| **IsA**$(x, y)$ | universal $x$ is a sub-kind of universal $y$ |
| **Precedes**$(x, y)$ | individual process $x$ precedes individual process $y$ |
| **HasParticipant**$(x, y)$ | individual $y$ participates in individual occurrent $x$ |
| **HasAgent**$(x, y)$ | individual $y$ is agent of individual occurrent $x$ |
| **Realizes**$(x, y)$ | individual process $x$ realizes individual function $y$ |
| **TypeOf**$(x, \mathbf{t}) = (x :: \mathbf{t})$ | individual $x$ is an object of type $\mathbf{t}$ |

Unlike in data-driven and statistical approaches where meaning is approximated by vectors in a multidimensional space, it is these primitive relations that for us form what we call 'dimensions of meaning'. In figure 4 we show how these relations are used to represent one meaning for the word 'book', namely "a written work or composition that has been published". Note that as an entity a 'book' (i) can be the agent of a changing event (as in 'Das Kapital **changed** many opinions over the years'); (ii) can have the popularity property (as in 'The Prince is a very **popular** book'); and (iii) can be the object of a writing event (as in 'William Shakespear **wrote** Hamlet').

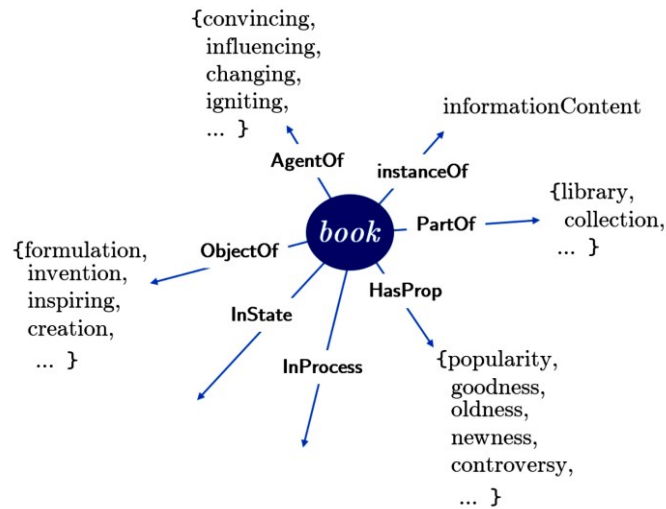

**Fig. 4.** Primitive and linguistically agnostic relations as the dimensions of word meaning.

Ironically, we can use GPT-4 to generate some of these vectors along the various dimensions since these systems are good at *predicting* a masked word. The data in figure 5 is obtained by asking GPT-4 to provide 25 ″plausible″ (or ″sensible″) replacements for the [**MASK**]. The three syntactic variations in the sentences are intended to recover entities along three dimensions, namely **agnetOf**, **objectOf** and **hasProp**. As we stated earlier this strategy can also be used to 'discover' the underlying ontology that seems to be implicit underneath our ordinary language. In figure 5 below we apply masking to generate the most plausible actions that a computer, a car, and a couch can be the object of. Note that while the three types of objects can be the objects of ASSEMBLE, we can sensibly say a computer or a car is RUNNING (or that a computer or a car is ON/OFF) but the same is not true of a couch. This tells us that while a computer and a car must have some common supertype (**machine**?), these two types seem to eventually belong to a different branch from couch although they all have a common supertype at some level of abstraction since they can all be 'assembled' (see figure 6).

Incidentally, defining meanings along these dimensions should also shed a new light on Wittgenstein's notion of ′family resemblance′ (see Aitchison, 2012). That is, while it might be difficult to think of a set of common features that define all kinds of *games*, *how we speak about all games* should not be different (see figure 7).

## 4    Concept Similarity

In figure 4 above we saw that a book can be described by the properties *influential* and *profound*, among others. That is, using the nominalized form we can say:

influence ∈ book . **hasProp**
profoundness ∈ book . **hasProp**

| *The book has* **[MASK]** *millions of people* | *Jon has* **[MASK]** *the book* | *Das Kapital was a very* **[MASK]** *book* |
|---|---|---|
| 1. influenced | 1. wrote | 1. influential |
| 2. inspired | 2. criticized | 2. impactful |
| 3. educated | 3. endorsed | 3. controversial |
| 4. affected | 4. debated | 4. challenging |
| 5. engaged | 5. read | 5. analytical |
| 6. perplexed | 6. quoted | 6. detailed |
| 7. challenged | 7. discussed | 7. scholarly |
| 8. reached | 8. interpreted | 8. complex |
| 9. enlightened | 9. appreciated | 9. profound |
| 10. motivated | 10. translated | 10. radical |
| 11. stirred | 11. reviewed | 11. enlightening |
| 12. provoked | 12. studied | 12. dense |
| 13. intrigued | 13. analyzed | 13. thought-provoking |
| 14. alarmed | 14. examined | 14. significant |
| 15. shaped | 15. dismissed | 15. rigorous |
| 16. guided | 16. understood | 16. comprehensive |
| 17. fascinated | 17. refuted | 17. polemical |
| 18. informed | 18. praised | 18. controversial |
| 19. captivated | 19. digested | 19. transformative |
| 20. provoked | 20. researched | 20. critical |
| 21. challenged | 21. referenced | 21. pivotal |
| 22. transformed | 22. challenged | 22. historical |
| 23. touched | 23. summarized | 23. theoretical |
| 24. awakened | 24. defended | 24. intricate |
| 25. stimulated | 25. bought | 25. philosophical |

**Fig. 5.** Querying GPT-4 to complete contexts with plausible actions/relations and properties that can plausibly (sensibly) be said of (or apply to) a book: as an agent of some action or activity, a book can influence, inspire, motivate, educate, etc. people; as the object of some activity, a book can be translated, interpreted, examined, refuted, etc. and finally, a book can have the property (or can be described as being) significant, critical, historical, influential, controversial, etc.

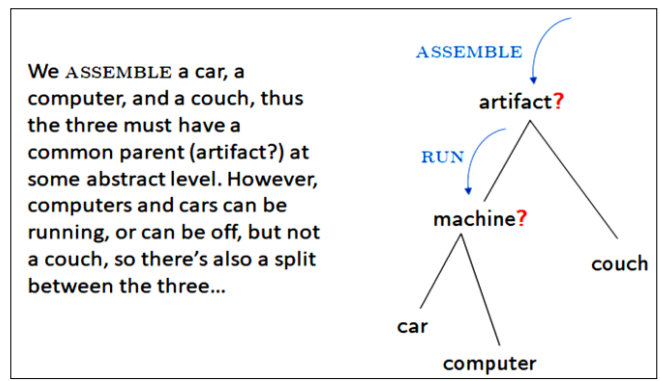

**Fig. 6.** A computer, a car, and a couch can be assembled, so at some level of abstraction they must have a common parent (artifact?). However, cars and computers, although not couches, RUN and can be described by being ON/OFF so they eventually must be in different branches.

| ObjectOf | AgentOf | HasProp |
|---|---|---|
| *John has* [MASK] *the game.* | *The **game** [MASK] everyone.* | *The **game** was very* [MASK]. |
| Won | Amazed | Exciting |
| Enjoyed | Thrilled | Difficult |
| Lost | Challenged | Enjoyable |
| Played | Engaged | Frustrating |
| Started | Entertained | Amazing |
| Finished | Frustrated | Innovative |
| Played | Bored | Disappointing |
| Abandoned | Exhausted | Entertaining |
| Dominated | Amused | Intriguing |
| Continued | Relaxed | Boring |
| Observed | Intrigued | Thrilling |
| Completed | Captivated | Challenging |
| Quit | Delighted | Easy |
| Dominated | Confused | Outstanding |
| Mastered | Bored | Complicated |

**Fig. 7.** While it is difficult to come up with features that are common to all *games*, we can uncover the descriptions we use in how we talk about games in everyday discourse.

Note that it might be the case that in our ordinary language use we speak of *influential* books more so than we speak of *profound* books. Thus, in general we might have

$(w_1, \text{influence}) \in \text{book} \,\textbf{.}\, \textbf{hasProp}$
$(w_2, \text{profoundness}) \in \text{book} \,\textbf{.}\, \textbf{hasProp}$

where $w_1 > w_2$ indicates that influence is used when describing books more than profoundness. The sets along some dimension $\textbf{D}_1$ therefore look like this:

$C_1 \,\textbf{.}\, \textbf{D}_1 = \{(w_{11}, p_{11}), (w_{12}, p_{12}), \dots \}$
$C_2 \,\textbf{.}\, \textbf{D}_1 = \{(w_{21}, p_{21}), (w_{22}, p_{22}), \dots \}$

The join for $C_1$ and $C_2$ along the dimension $\textbf{D}_1$ is then computed as follows:

$\textsc{DimensionJoin}(C_1, C_2, \textbf{D}_1)$
$= \{\langle (w_1, p_1), (w_2, p_2) \rangle \mid (p_1 = p_2) \wedge (w_1, p_1) \in C_1 \,\textbf{.}\, \textbf{D}_1 \wedge (w_2, p_2) \in C_2 \,\textbf{.}\, \textbf{D}_1\}$

As an example, we might have the following join along the **hasProp** dimension for one meaning of 'book' and one meaning for 'publication':

$\textsc{DimensionJoin}(book_1, publication_3, \textbf{hasProp})$
$= \{\langle (0.75, \text{popularity}), (0.72, \text{popularity}) \rangle,$
$\quad \langle (0.73, \text{controversy}), (0.71, \text{controversy}) \rangle, \dots \}$

The similarity along the **hasProp** dimension can now be computed as follows:

$\textsc{DimensionJoin}(book_1, publication_3, \textbf{hasProp})$
$= \textbf{sum}([\textsc{FeatureSim}(p) \text{ for p in } fs]) \,/\, |fs|$
$\qquad \text{where}$
$\qquad\qquad fs = \textsc{FeatureSet}(book_1, publication_3, \textbf{hasProp})$
$\qquad\qquad \textsc{FeatureSim}\langle (w_1, p_1), (w_2, p_2) \rangle = \textbf{if } p_1 = p_2 \textbf{ then } 1 - \textbf{abs}(w_1 - w_2) \textbf{ else } 0$

The final similarity between $book_1$ and $publication_3$ is then a weighted average of the similarity across all dimensions, where, for simplicity, the weights are all equal here:

CSIMILARITY($book_1$, $publication_3$)
= **sum**([DSIMILARITY($book_1$, $publication_3$, **D**) for **D** in $dims$]) / |$dims$|
    where
       $dims$ = {**hasProp**, **agentOf**, **objectOf**, **inState**, **partOf**, … }

One final note regarding concept similarity is that the above similarity is based on linguistic dimensions – that is, it is a similarity based on how we sensibly speak about concepts in our ordinary language. Thus, it will be expected that a *book* and a *publication*, for example, are quite similar, and not because their vectors are in close proximity in a multidimensional space, but (crucially) because almost anything that can sensibly be said of a book can also be said of a publication, and vice versa. The implications of this change of perspective are substantial. While current LLMs can only account for proximity and similarity, they do not account for concept 'identity' which makes their concept representation circular as has been eloquently pointed out by (Lopes, 2023).

## 5      Concluding Remarks

While LLMs have shown impressive capabilities in producing coherent and human-like text, these models do not truly 'understand' language and they do not differentiate between factual and nonfactual information. LLMs will also fail in making correct inferences in intensional contexts or in contexts where syntactic patterns with high probabilities must be overruled by semantics and background knowledge. Since it is our firm belief that the relative success of LLMs is due to a successful bottom-up reverse engineering strategy, we suggested in this paper applying this method in a symbolic setting. The reverse-engineering method we proposed allows us also to 'discover' the ontology that seems to be implicit underneath language. The reader is referred to (Saba, 2020) for a detailed discussion on how the system described here can be used to handle most challenges in the semantics of natural language.

## References

1. Aitchison, J.: *Words in the Mind – An Introduction to the Mental Lexicon*, Wiley (2012)
2. Asher, N.: *Lexical Meaning in Context, a Web of Words*. Cambridge University Press (2011)
3. Asher, N. & Pustejovsky, J.: A type composition logic for generative lexicon, *Journal of Cognitive Science* 6: 1–38 (2011)
4. Boleda, G.: Distributional Semantics and Linguistic Theory, *Annual Review of Linguistics*, 6, pp. 213–34 (2020)
5. Dummett, M.: *Frege: Philosophy of Language*. Harvard University Press (1981)
6. Fodor, J. A. and Pylyshyn, Zenon W.: Connectionism and cognitive architecture: A critical analysis, Cognition, 28 (1), pp. 3-71 (1988)
7. Hobbs, J.: Ontological promiscuity. In *Proc. of the 23rd Annual Meeting of the Assoc. for Computational Linguistics*, Chicago, Illinois, 1985, pp. 61–69 (1985)
8. Harris, Z. S.: Distributional Structure. *Word* 10:146–62 (1954)

9. Kiss, K. E. and Pafel, J.: Quantifier Scope Ambiguities, In Martin Everaert and Henk C. van Riemsdijk (Eds.), *The Wiley Blackwell Companion to Syntax* (2017)

10. Larson, R.: Events and Modification in Nominals, In Devon Strolovitch and Aaron Lawson (eds), *SALT VIII*, 145- 168, Ithaca, NY (1998)

11. Lidz, J.: Children's Use of Syntax in Word Learning, In Anna Papafragou, John C. Trueswell & Lila R. Gleitman (eds.), *The Oxford Handbook of the Mental Lexicon*, Oxford University Press (2022)

12. Lopes, J.: Can Deep CNNs Avoid Infnite Regress/Circularity in Content Constitution?, Minds and Machines, https://doi.org/10.1007/s11023-023-09642-0 (2023)

13. McShane, M., Beale, S. and Babkin, P.: Nominal Compound Interpretation by Intelligent Agents, *Linguistic Issues in Language Technology* (*LiLT*), vol. 10, No 1 (2014)

14. Milne, P.: Frege's Context Principle, *Mind*, Vol. **95**, No. 380, pp. 491-495 (1986)

15. Moltmann, F.: *Abstract Objects and the Semantics of Natural Language*, Oxford University Press (2013)

16. Peckenpaugh, T.: *Prepositional phrase attachment ambiguities in declarative and interrogative contexts: Oral reading data*, PhD Thesis, The City University of New York (2019)

17. Piñango, M. M., Zhang, M., et. al., Metonymy as Referential Dependency: Psycholinguistic and Neurolinguistic Arguments for a Unified Linguistic Treatment, *Cognitive Science*, 41 (S2), pp. 351–378 (2017)

18. Saba, W.: New Research Vindicates Fodor and Pylyshyn: No Explainable AI Without 'Structured Semantics, *Blog of the Communications of the ACM*, September 14 (2022)

19. Saba, W.: Language, Knowledge and Ontology: Where Formal Semantics Went Wrong, and How to Go Forward, Again, *Journal of Knowledge Structures and Systems* (JKSS), **1** (1), pp. 40-62 (2020)

20. Saba, W. and Corriveau, J.-P.: Plausible Reasoning and the Resolution of Quantifier Scope Ambiguities, *Studia Logica – Int. Journal of Symbolic Logic*, 67, pp. 271-289. (2001)

21. Saba, W.: Language, logic and ontology: Uncovering the structure of commonsense knowledge, *Int. J. of Human Computer Studies*, **7**(65), pp. 610-623 (2007)

22. Shelestiuk, H. V.: Metonymy as a tool of cognition and representation: A natural language analysis, *Semiotica*, pp. 1-20 (2005)

23. Sommers, F.: Types and ontology. *Philosophical Review*, **72** (3), 327-363 (1963)

24. Sugawara, S. and Tsugita. S.: On Degrees of Freedom in Defining and Testing Natural Language Understanding, In *Findings of the Association for Computational Linguistics: ACL*, pp. 13625–13649 (2023)

25. Vaswani, A., Shazeer, N., et. al.: Attention is All You Need, In *NIPS'17: Proceedings of the 31st Int. Conference on Neural Information Processing Systems*. pp. 6000–6010, (2017)

26. Viebahn, E.: Copredication, polysemy and context-sensitivity, *Inquiry*, Volume 65 (2020)

27. von Fintel, K. and Heim, I. *Lecture Notes on Intensional Semantics*, available online here https://www.phil-fak.uni-duesseldorf.de/summerschool2002/fintel.pdf, (2002)